\documentclass[runningheads]{llncs}
\usepackage{graphicx}
\usepackage{multirow}
\usepackage{booktabs}
\usepackage{bbding}
\usepackage{hyperref, color}
\usepackage{amsmath}
\usepackage{algorithm}
\usepackage{algorithmic}
\usepackage[switch]{lineno}
\usepackage{threeparttable}
\usepackage{rotating}

\begin{document}
\title{SAM 2 in Robotic Surgery: An Empirical Evaluation for Robustness and Generalization in Surgical Video Segmentation}
\titlerunning{SAM 2 in Robotic Surgery}
% If the paper title is too long for the running head, you can set
% an abbreviated paper title here
%
% \author{Anonymous Authors}
\author{Jieming Yu\inst{1,2,3}
\and {An Wang}\inst{1,2}
\and {Wenzhen Dong}\inst{1,2}
\and Mengya Xu\inst{1,3}
\and Mobarakol Islam\inst{4}
\and Jie Wang\inst{2}
\and Long Bai\inst{1,2}
\thanks{Project Lead.}
\and Hongliang Ren\inst{1,2} 
\thanks{Corresponding Author.}
}

\authorrunning{J. Yu et al.}
% First names are abbreviated in the running head.
% If there are more than two authors, 'et al.' is used.
%
\institute{Dept. of Electronic Engineering, The Chinese University of Hong Kong (CUHK), Hong Kong SAR, China
\and Shenzhen Research Institute, CUHK, Shenzhen, China
\and Dept. of Computer Science and Engineering, CUHK, Hong Kong SAR, China
\and Dept. of Medical Physics and Biomedical Engineering, Wellcome/EPSRC Centre for Interventional and Surgical Sciences, University College London, London, UK\\
\email{ireneyu1024@gmail.com, b.long@link.cuhk.edu.hk, hlren@ee.cuhk.edu.hk}
}
\maketitle              % typeset the header of the contribution
\begin{abstract}
The recent Segment Anything Model (SAM) 2 has demonstrated remarkable foundational competence in semantic segmentation, with its memory mechanism and mask decoder further addressing challenges in video tracking and object occlusion, thereby achieving superior results in interactive segmentation for both images and videos. Building upon our previous empirical studies, we further explore the zero-shot segmentation performance of SAM 2 in robot-assisted surgery based on prompts, alongside its robustness against real-world corruption. For static images, we employ two forms of prompts: 1-point and bounding box, while for video sequences, the 1-point prompt is applied to the initial frame. Through extensive experimentation on the MICCAI EndoVis 2017 and EndoVis 2018 benchmarks, SAM 2, when utilizing bounding box prompts, outperforms state-of-the-art (SOTA) methods in comparative evaluations. The results with point prompts also exhibit a substantial enhancement over SAM's capabilities, nearing or even surpassing existing unprompted SOTA methodologies. Besides, SAM 2 demonstrates improved inference speed and less performance degradation against various image corruption. Although slightly unsatisfactory results remain in specific edges or regions, SAM 2's robust adaptability to 1-point prompts underscores its potential for downstream surgical tasks with limited prompt requirements.
% \keywords{}
\end{abstract}
\section{Introduction}
\label{sec:introduction}

%%%%%%%%%%%%%%%%%% Intro %%%%%%%%%%%%%%%%%%%%%%%%%%%%

Surgical instrument segmenting and tracking is a significant topic, with its rich representation contributing to the development of various downstream applications~\cite{baste2018development,pakhomov2020searching,wang2023rethinking}. Unfortunately, due to a lack of large-scale surgical data, the efforts toward surgical foundational models are significantly lagging compared to general computer vision. Specifically, the acquisition and annotation of high-quality surgical data require expensive resources and human labor, while synthetic data training falls short of distribution diversity and real-world applications.

The segmentation foundation models, which are trained on more than one billion masks, have made great progress in the field of natural image segmentation, but tend to fail in common medical scenarios because of the large domain gap~\cite{deng2023segment,he2023accuracy,hu2023sam,ma2023segment}. To tackle this issue, researchers have adapted Segmentation Attention Models (SAM)~\cite{kirillov2023segment} from general 2D vision to medical applications, capitalizing on the rich, diverse pre-trained data through Parameter-Efficient Fine-Tuning (PEFT) techniques such as adapters or low-rank adaptations (LoRA)~\cite{hu2021lora}. For example, the Med-SAM-Adapter~\cite{wu2023medical} leverages medical-specific domain knowledge to refine segmentation models effectively, via a simple yet effective adapter. Similarly, SAMed~\cite{zhang2023customized} employs a low-rank finetuning strategy on both the image encoder and the prompt encoder, alongside the mask decoder, for medical image segmentation tasks. Furthermore, a series of works have been proposed for automating prompts and fine-tuning SAM models to perform end-to-end semantic segmentation tasks in surgical scenarios, including two-stage strategy~\cite{yu2024adapting}, text prompt~\cite{paranjape2024adaptivesam}, automating bounding box prompt~\cite{sheng2024surgical}, and feature matching~\cite{murali2024cyclesam}.

The recently introduced SAM 2~\cite{ravi2024sam}, leveraging the unique memory mechanism and mask decoder, has demonstrated notable superiority in effectiveness and efficiency over its previous version. SAM 2 successfully manages complex scenarios characterized by detailed anatomical structures, motion, and occlusion, thereby enhancing the model's reliability across a more extensive array of applications. Benefiting from SAM 2's robust capability in handling complex scenarios, it demonstrates significant potential in processing surgical scene data.
Consequently, it is necessary to evaluate the performance of SAM 2 in medical contexts and its robustness under real-world corruption for further investigation.

Following our previous work on SAM~\cite{wang2023sam}\footnote{\url{https://arxiv.org/abs/2308.07156}}, we assess the generalizability of SAM 2 across various operational scenarios. Specifically, our contributions and findings can be summarized as:
\begin{itemize}
    \item We conduct a comprehensive empirical study on surgical images and videos based on SAM 2. For surgical images, we use the bounding box or 1-point as the prompt input; for videos, we prompt 1 point in the first frame. The results indicate that SAM 2 overall outperforms its previous version. 
    \item By using bounding boxes as prompts, SAM 2 has become the new state-of-the-art (SOTA) in the surgical domain. Surprisingly, in video segmentation, SAM 2, which applies the 1-point prompt only in the first frame, demonstrated superior performance compared to SAM with images using the 1-point prompt for every frame. 
    \item We further evaluate the robustness of SAM 2 by analyzing its performance on synthetic surgical datasets, which include diverse levels of corruption and perturbations. With bounding boxes as prompts, SAM 2 on images exhibited robust resistance to real-world corruption, showing minimal performance degradation when faced with challenges such as image compression, noise, blur, and occlusion.
\end{itemize}

%%%%%%%%%%%%%%%%%%%%%%%%%%%%%%%%%%%%%%%%%%%%%%
\begin{table}[h]
  \centering
  \caption{Quantitative comparison of binary and instrument segmentation on EndoVis17 and EndoVis18 datasets. For SAM 2, we present the results in images and videos.}
  \begin{threeparttable}
    \resizebox{0.95\textwidth}{!}{
    \begin{tabular}{cccccccc}
    \toprule
    \multirow{2}[2]{*}{Type} & \multirow{2}[2]{*}{Method} & \multirow{2}[2]{*}{Pub/Year(20-)} & \multirow{2}[2]{*}{Arch.} & \multicolumn{2}{c}{EndoVis17} & \multicolumn{2}{c}{EndoVis18} \\
\cmidrule{5-8}          &       &       &       & Binary IoU & Instrument IoU & Binary IoU & Instrument IoU \\
    \midrule
    \multirow{5}[2]{*}{Single-Task} & Vanilla UNet & MICCAI15 & UNet  & 75.44 & 15.80 & 68.89 & - \\
          & TernausNet & ICMLA18 & UNet  & 83.60 & 35.27 & -     & 46.22 \\
          & MF-TAPNet & MICCAI19 & UNet  & 87.56 & 37.35 & -     & 67.87 \\
          & Islam et al. & RA-L19 & -     & 84.50 & -     & -     & - \\
          & ISINet & MICCAI21 & Res50 & -     & 55.62 & -     & 73.03 \\
          & Wang et al. & MICCAI22 & UNet  & -     & -     & 58.12 & - \\
    \midrule
    \multirow{5}[0]{*}{Multi-Task} & ST-MTL & MedIA21 & -     & 83.49 & -     & -     & - \\
          & AP-MTL & ICRA20 & -     & 88.75 & -     & -     & - \\
          & S-MTL & RA-L22 & -     & -     & -     & -     & 43.54 \\
          & TraSeTR & ICRA22 & Res50 + Trfm & -     & 60.40 & - & 76.20 \\
          & S3Net & WACV23 & Res50 & -     & 72.54 & -     & 75.81 \\
    \midrule
    \multirow{6}[0]{*}{Prompt-based} & SAM (1 Point) & arxiv23 & ViT\_h   & 53.88 & 55.96\tnote{*} & 57.12 & 54.30\tnote{*} \\
          & SAM (Box) & arxiv23 & ViT\_h   & 89.19 & \textbf{88.20}\tnote{*} & 89.35 & 81.09\tnote{*} \\
          & SAM 2-Image (1 Point) & arxiv24 & ViT\_h  &84.96 & 81.10\tnote{*} & 77.14 &73.76\tnote{*} \\
          & SAM 2-Image (Box)  & arxiv24 & ViT\_h  & \textbf{90.97} & 86.92\tnote{*} & \textbf{90.18} & \textbf{81.97}\tnote{*}  \\
          & SAM 2-Video (1 Point) & arxiv24 & ViT\_h &62.45  &58.74\tnote{*} &65.19&57.59\tnote{*} \\
          % & SAM 2 Box/Video & arxiv24 & ViT\_h    \\
    \bottomrule
    \end{tabular}%
    }\end{threeparttable}
    \begin{tablenotes}
    \footnotesize
    \item{*} Categorical information directly inherits from associated prompts.
    \end{tablenotes}
  \label{tab:overall_res}%
\end{table}%

%%%%%%%%%%%%%%%%%%%%% Figure %%%%%%%%%%%%%%%%%%%%%
\begin{figure*}[htb]
  \centering
  \includegraphics[width=0.8\linewidth]{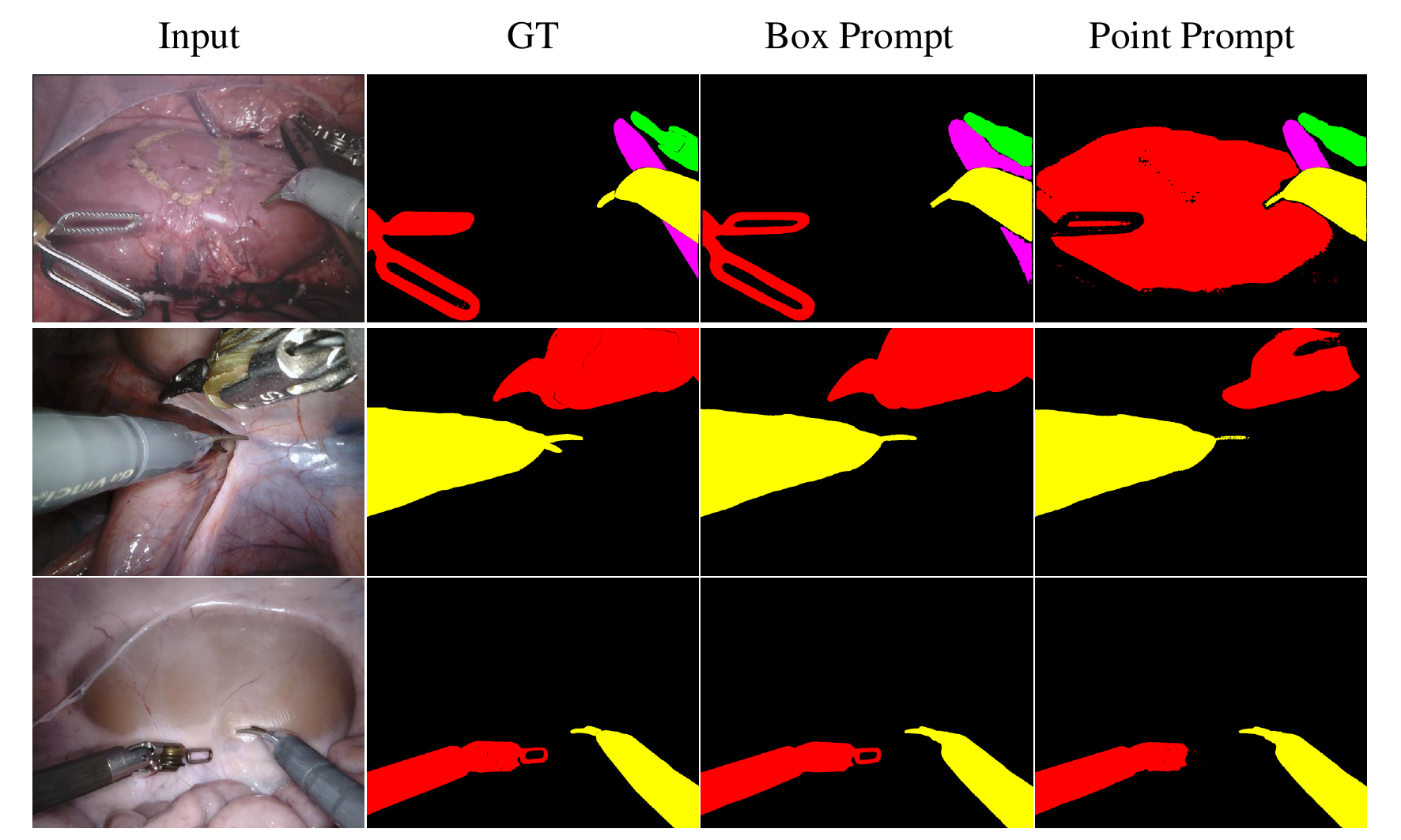}
  \caption{Qualitative results of SAM 2 on three images of the surgical scene.}
  \label{fig:quali}
\end{figure*}
%%%%%%%%%%%%%%%%%%%%% Figure %%%%%%%%%%%%%%%%%%%%%

\begin{table}
    \centering
    \caption{Comparison of inference speed (fps) of SAM and SAM 2 on EndoVis18~\cite{allan20202018}. Experiments are conducted on 1 RTX 3090 GPU, with Pytorch 2.4 and CUDA 12.1.}
    \label{tab:fps}
    \begin{tabular}{c|cc}
    \toprule
        Model & Point & Box \\
    \hline
        SAM & 2.95 & 3.04\\
        SAM 2 & 8.75 & 8.94\\
    \bottomrule
    \end{tabular}
\end{table}

\section{Surgical Instruments Segmentation with Prompts}
\subsubsection{Implementation}
MICCAI EndoVis17~\cite{allan20192017} and EndoVis18~\cite{allan20202018} are used for our evaluation and we follow all the validation set splitting in~\cite{wang2023sam}.  For images, we utilize either a single point or a bounding box as prompts. For videos, we employ a single point from the initial frame as the prompt. The bounding box originates from manual annotations in our previous work~\cite{bai2023surgical}, while the single point is derived by calculating the centroid of the corresponding mask. Since the output of SAM 2 does not contain class information, we directly assign the class information from the input prompt to the output mask to prevent poor performance caused by incorrect class information.

\subsubsection{Comparison methods}
We follow the comparison methods in our previous version as follows: vanilla UNet~\cite{ronneberger2015u}, TernausNet~\cite{shvets2018automatic}, MF-TAPNet~\cite{jin2019incorporating}, Islam et al.~\cite{islam2019real}, Wang et al.~\cite{wang2022rethinking}, ST-MTL~\cite{islam2021st}, S-MTL~\cite{seenivasan2022global}, AP-MTL~\cite{islam2020ap}, ISINet~\cite{gonzalez2020isinet}, TraSeTR~\cite{zhao2022trasetr}, and S3Net~\cite{baby2023forks} for surgical binary and instrument-wise segmentation. The SAM results are adopted from~\cite{wang2023sam}, and we use the SAM 2-Hiera-Large~\cite{kirillov2023segment} for our evaluation. It is important to note that a completely fair comparison cannot be achieved, as existing methods do not require prompts during inference.

\subsubsection{Results and Analysis}
Overall, SAM~2~\cite{ravi2024sam} demonstrates better performance compared to SAM~\cite{kirillov2023segment} in Table~\ref{tab:overall_res} and~\ref{tab:fps}. The results using the bounding box prompt maintain SOTA performance over previous unprompted methods, but its improvement on SAM~\cite{kirillov2023segment} is not significant. In terms of the 1-point prompt, SAM~2~\cite{ravi2024sam} exhibits substantial enhancements, with overall performance increasing by 20\%–30\%. Furthermore, SAM~2~\cite{ravi2024sam} possesses an inference speed more than twice that of SAM~\cite{kirillov2023segment}, greatly benefiting practical clinical applications; doctors can simply click, and the model quickly provides results for the corresponding targets in the images. 

Additionally, in SAM 2, we compare the differences in the segmentation results between video and image when using a 1-point prompt. The performance of video segmentation, with only the first frame prompt, is lower than that of image segmentation. The results are consistent with the results presented in the SAM 2 technical report, since the image segmentation task will get more prompt information – the 1-point prompt will be applied at each frame. 
However, despite this, the video segmentation results with the 1-point prompt in SAM 2 still outperform the previous SAM image segmentation results using the 1-point prompt, enabling satisfactory results for downstream applications.

% Table generated by Excel2LaTeX from sheet 'Sheet3'
\begin{table}[!t]
  \centering
  \caption{Quantitative results on various corrupted EndoVis18 validation data.The prompts used by SAM-Image Segmentation, SAM 2-Image Segmentation, and SAM 2-Video Segmentation are bounding box, bounding box, and point, respectively.}
    \resizebox{\textwidth}{!}{
    \begin{tabular}{c|c|c|cccc|ccccc|ccccc|cccc}
    \toprule
    
    \multirow{2}[0]{*}{Task} & & \multirow{2}[0]{*}{Severity} & \multicolumn{4}{c|}{Noise}    & \multicolumn{5}{c|}{Blur}             & \multicolumn{5}{c|}{Weather}          & \multicolumn{4}{c}{Digital} \\
          &        &       & Gaussian & Shot  & Impulse & Speckle & Defocus & Glass & Motion & Zoom  & Gaussian & Snow  & Frost   & Fog     & Bright  & Spatter & Contrast & Pixel & JPEG  & Saturate \\
    \hline
    \multirow{12}[2]{*}{\begin{sideways}SAM - Image Segmentation\end{sideways}} & \multirow{6}[2]{*}{\begin{sideways}Binary\end{sideways}} & \multicolumn{1}{c}{0} & \multicolumn{18}{|c}{89.35} \\\cline{3-21}          
    & & 1     & 77.69 & 80.18 & 80.43 & 83.28 & 82.01 & 80.53 & 82.99 & 80.30 & 85.40 & 84.08 & 83.12 & 85.38 & 87.43 & 86.69 & 85.76 & 81.12 & 58.77 & 86.64 \\
    & & 2     & 73.92 & 76.07 & 76.15 & 81.65 & 80.21 & 79.20 & 80.22 & 77.55 & 81.69 & 80.69 & 80.34 & 84.65 & 87.27 & 84.21 & 84.90 & 79.32 & 56.04 & 84.85 \\
    & & 3     & 69.21 & 71.74 & 73.02 & 77.74 & 76.96 & 72.64 & 75.50 & 75.27 & 78.31 & 79.58 & 78.90 & 83.62 & 87.23 & 82.50 & 83.36 & 73.81 & 56.25 & 86.84 \\
    & & 4     & 63.80 & 65.41 & 67.29 & 75.28 & 73.79 & 72.38 & 69.60 & 73.22 & 75.23 & 76.33 & 78.38 & 82.28 & 87.06 & 83.12 & 77.12 & 70.82 & 57.59 & 83.21 \\
    & & 5     & 57.07 & 60.61 & 61.61 & 71.83 & 69.85 & 69.59 & 66.25 & 71.58 & 66.96 & 77.66 & 76.82 & 78.84 & 86.43 & 79.62 & 66.58 & 68.55 & 56.77 & 81.26 \\
    \cline{2-21}
    & \multirow{6}[2]{*}{\begin{sideways}Instrument\end{sideways}} & \multicolumn{1}{c}{0} & \multicolumn{18}{|c}{81.09} \\\cline{3-21}           
    & & 1     & 69.51 & 71.83 & 72.25 & 74.82 & 73.64 & 72.13 & 74.33 & 71.41 & 76.79 & 75.40 & 74.42 & 76.82 & 79.16 & 78.24 & 77.17 & 72.94 & 54.86 & 78.27 \\
    & & 2     & 66.06 & 68.09 & 68.53 & 73.19 & 71.74 & 71.02 & 71.46 & 68.85 & 73.15 & 72.13 & 71.65 & 76.14 & 79.00 & 75.54 & 76.22 & 71.55 & 52.23 & 76.61 \\
    & & 3     & 62.01 & 64.44 & 65.89 & 69.75 & 68.74 & 64.97 & 67.13 & 67.12 & 70.08 & 70.97 & 70.21 & 75.01 & 78.90 & 73.70 & 74.67 & 66.83 & 51.63 & 78.39 \\
    & & 4     & 57.28 & 59.12 & 61.03 & 67.82 & 65.87 & 64.87 & 62.15 & 65.18 & 67.23 & 68.43 & 69.79 & 73.73 & 78.73 & 74.24 & 69.48 & 63.99 & 51.88 & 74.91 \\
    & & 5     & 51.56 & 55.16 & 55.86 & 64.76 & 62.43 & 62.23 & 59.26 & 63.96 & 60.60 & 69.33 & 68.32 & 70.45 & 78.19 & 70.72 & 61.14 & 61.79 & 51.01 & 73.35 \\
    \hline
    \multirow{12}[2]{*}{\begin{sideways}SAM 2 - Image Segmentation\end{sideways}} & \multirow{6}[2]{*}{\begin{sideways}Binary\end{sideways}} & \multicolumn{1}{c}{0} & \multicolumn{18}{|c}{90.18} \\\cline{3-21} 
    & & 1     & 85.20 & 86.24 & 85.20 & 87.69 & 85.53 & 84.95 & 85.46 & 81.83 & 87.34 & 87.81 & 89.50 & 87.36 & 88.96 & 88.63 & 87.79 & 86.92 & 85.68 & 88.80 \\
    & & 2     & 82.04 & 83.69 & 82.38 & 86.77 & 83.93 & 84.43 & 82.71 & 79.18 & 85.39 & 85.06 & 89.48 & 86.88 & 88.96 & 87.78 & 87.31 & 86.39 & 83.58 & 88.01 \\
    & & 3     & 77.35 & 80.01 & 80.19 & 83.94 & 81.56 & 79.16 & 78.72 & 77.12 & 83.09 & 85.20 & 89.43 & 85.34 & 88.90 & 87.13 & 86.13 & 83.24 & 82.33 & 88.54 \\
    & & 4     & 72.75 & 74.65 & 75.99 & 81.91 & 79.31 & 79.41 & 74.01 & 75.24 & 80.75 & 83.30 & 89.42 & 84.05 & 88.90 & 87.23 & 82.08 & 80.75 & 74.81 & 86.61 \\
    & & 5     & 68.09 & 71.30 & 71.47 & 79.56 & 77.06 & 78.07 & 71.52 & 73.40 & 76.55 & 83.04 & 89.46 & 80.70 & 88.61 & 85.28 & 72.93 & 78.78 & 65.16 & 84.96 \\
    \cline{2-21}
    & \multirow{6}[2]{*}{\begin{sideways}Instrument\end{sideways}} & \multicolumn{1}{c}{0} & \multicolumn{18}{|c}{81.97} \\\cline{3-21} 
    & & 1     & 76.61 & 77.56 & 76.58 & 79.24 & 76.99 & 76.47 & 76.95 & 72.99 & 78.96 & 79.33 & 81.21 & 78.93 & 80.73 & 80.39 & 79.39 & 78.39 & 77.00 & 80.50 \\
    & & 2     & 73.76 & 75.25 & 74.19 & 78.29 & 75.37 & 75.97 & 74.21 & 70.30 & 76.69 & 76.40 & 81.12 & 78.20 & 80.75 & 79.56 & 78.71 & 77.88 & 74.91 & 79.62 \\
    & & 3     & 69.65 & 72.02 & 72.20 & 75.57 & 72.85 & 70.72 & 70.14 & 68.64 & 74.25 & 76.74 & 81.00 & 76.72 & 80.73 & 78.76 & 77.44 & 74.78 & 73.72 & 80.39 \\
    & & 4     & 65.68 & 67.47 & 68.61 & 73.97 & 70.63 & 70.98 & 65.88 & 67.00 & 72.04 & 74.71 & 81.00 & 75.61 & 80.79 & 78.75 & 73.72 & 72.39 & 67.18 & 78.45 \\
    & & 5     & 61.44 & 64.25 & 64.34 & 71.99 & 68.53 & 69.52 & 63.56 & 65.70 & 68.24 & 74.76 & 81.04 & 72.62 & 80.55 & 76.66 & 65.79 & 70.83 & 58.93 & 77.10 \\
    \hline
    \multirow{12}[2]{*}{\begin{sideways}SAM 2 - Video Segmentation\end{sideways}} & \multirow{6}[2]{*}{\begin{sideways}Binary\end{sideways}} & \multicolumn{1}{c}{0} & \multicolumn{18}{|c}{65.19} \\\cline{3-21} 
    & & 1 & 60.36 & 58.86 & 23.57 & 63.22 & 55.76 & 71.84 & 60.07 & 53.31 & 62.74 & 50.52 & 70.92 & 58.47 & 65.10 & 72.62 & 62.87 & 60.63 & 57.53 & 64.71 \\
    & & 2 & 43.53 & 60.07 & 26.01 & 54.63 & 52.33 & 64.71 & 50.44 & 64.93 & 55.77 & 45.60 & 71.82 & 66.17 & 72.26 & 68.72 & 63.42 & 66.21 & 49.61 & 63.37 \\
    & & 3 & 31.64 & 40.17 & 34.03 & 52.60 & 48.08 & 57.57 & 47.44 & 60.00 & 53.22 & 43.29 & 69.59 & 59.88 & 70.37 & 65.78 & 51.68 & 42.22 & 51.51 & 68.40 \\
    & & 4 & 20.54 & 23.87 & 30.87 & 46.71 & 57.18 & 61.08 & 53.64 & 58.31 & 50.67 & 44.38 & 70.52 & 56.90 & 72.30 & 58.02 & 50.86 & 42.24 & 47.94 & 45.04 \\
    & & 5 & 18.45 & 18.73 & 21.75 & 33.14 & 55.21 & 57.92 & 46.85 & 57.40 & 34.01 & 43.38 & 70.42 & 38.73 & 64.67 & 60.73 & 25.44 & 37.81 & 39.39 & 57.04 \\
    \cline{2-21}
    & \multirow{6}[2]{*}{\begin{sideways}Instrument\end{sideways}} & \multicolumn{1}{c}{0} & \multicolumn{18}{|c}{57.59} \\\cline{3-21} 
    & & 1 & 48.09 & 50.67 & 19.96 & 56.22 & 42.55 & 58.50 & 47.62 & 41.00 & 49.44 & 43.92 & 58.12 & 46.45 & 56.92 & 58.80 & 52.42 & 54.80 & 46.13 & 55.64 \\
    & & 2 & 37.18 & 49.89 & 20.43 & 48.89 & 40.56 & 41.95 & 44.26 & 42.45 & 44.20 & 38.90 & 57.69 & 54.17 & 62.65 & 58.77 & 50.86 & 60.96 & 38.71 & 52.41 \\
    & & 3 & 28.82 & 37.78 & 29.80 & 41.97 & 38.73 & 47.11 & 46.29 & 45.92 & 40.59 & 31.40 & 55.53 & 48.24 & 56.10 & 54.22 & 40.19 & 42.59 & 43.53 & 58.02 \\
    & & 4 & 15.47 & 16.06  & 24.24 & 36.74 & 45.70 & 48.94 & 43.36 & 44.94 & 35.05 & 37.68 & 56.61 & 40.14 & 57.63 & 52.65 & 36.11 & 30.11 & 34.76 & 39.73 \\
    & & 5 & 9.30 & 13.21 & 15.60 & 26.89 & 40.97 & 45.50 & 36.27 & 39.54 & 19.01 & 37.36 & 56.22 & 29.38 & 51.90 & 54.11 & 12.77 & 24.34 & 27.43 & 46.64 \\
    \bottomrule
    \end{tabular}%
    }
  \label{tab:corruption}%
\end{table}%

\section{Robustness under Data Corruption}
\subsubsection{Implementation}

We introduce image perturbations to evaluate robustness against input variations and analyze performance discrepancies. According to the robustness evaluation benchmark~\cite{hendrycks2018benchmarking}, SAM~\cite{kirillov2023segment} and SAM 2~\cite{ravi2024sam} underwent assessment across 18 types of data corruptions spanning 5 severity levels, following the official implementations\footnote{\url{https://github.com/hendrycks/robustness}}. Specifically, these data corruptions are (i) \textit{Blur} (defocus, glass, motion, zoom, Gaussian); (ii) \textit{Digital} (contrast, pixel, jpeg); and (iii) \textit{Noise} (Gaussian, Shot, Impulse, Speckle); (iv) \textit{Weather} (snow, frost, fog, brightness); (v) Others (spatter, saturate). The exclusion of the \textit{Elastic Transformation} was necessary to ensure alignment between input images and their corresponding masks.

% %%%%%%%%%%%%%%%%%%%%% Figure %%%%%%%%%%%%%%%%%%%%%
\begin{figure*}[!t]
  \centering
  \includegraphics[width=0.8\linewidth]{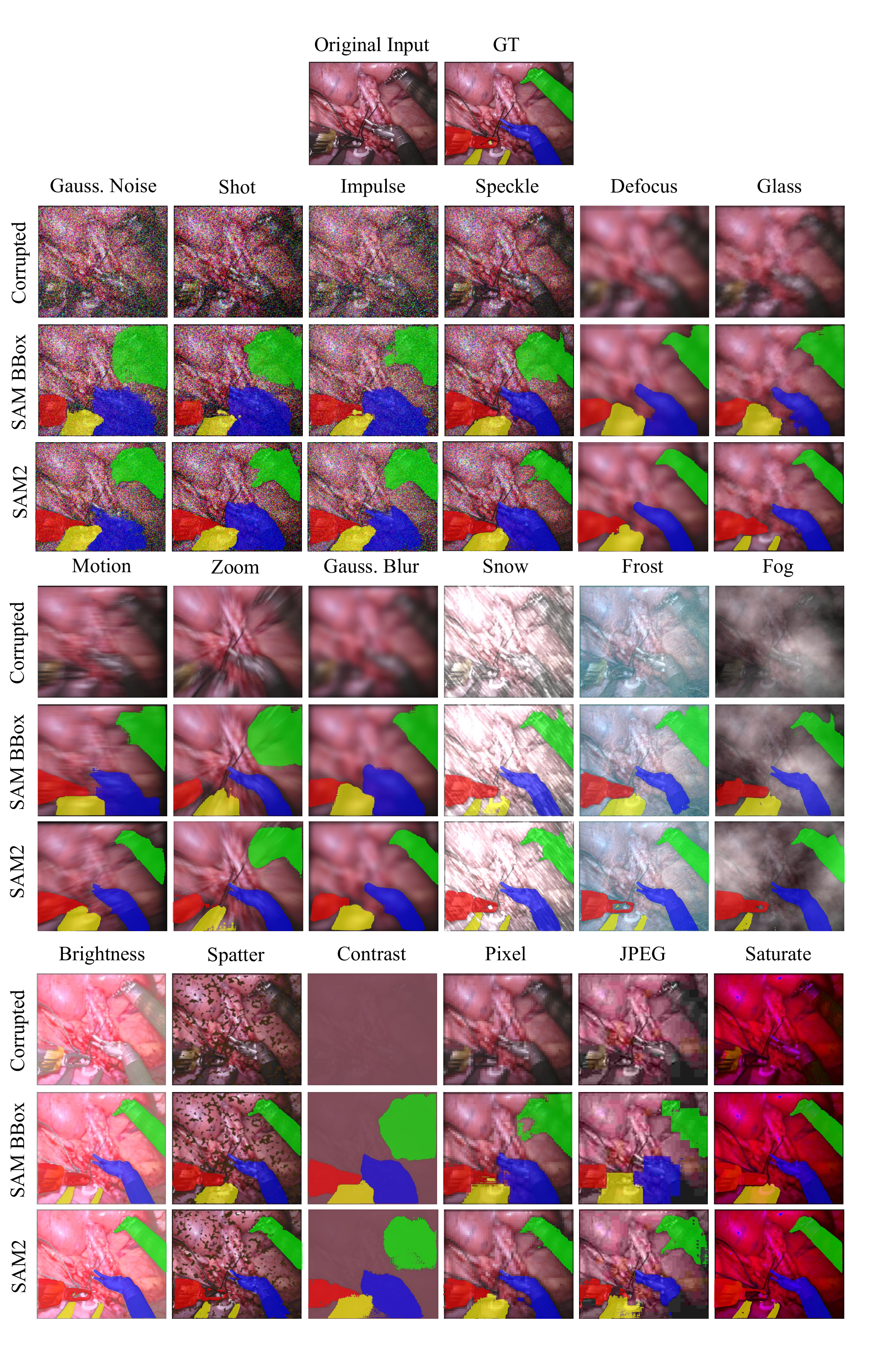}
  \caption{Qualitative results of SAM 2 under 18 data corruptions of level-5 severity. Given that the implementation of specific transformations (e.g., spatter) relies on random functions, and the corrupted dataset in our previous version is no longer accessible, we have regenerated the corrupted images. While some types of images may exhibit slight variations, the overall statistical consistency ensures the reliability of our findings.}
  \label{fig:corrupt}
\end{figure*}
% %%%%%%%%%%%%%%%%%%%%% Figure %%%%%%%%%%%%%%%%%%%%%

\subsubsection{Results and Analysis}

The extent of data corruption correlates directly with the degradation in the performance of SAM~\cite{kirillov2023segment} and SAM 2~\cite{ravi2024sam}, as illustrated in Table~\ref{tab:corruption}. SAM-Image Segmentation and SAM 2-Image Segmentation use bounding box prompts. Considering that SAM 2~\cite{ravi2024sam} does not offer bounding boxes as the prompt interface, SAM 2-Video Segmentation employs point prompts. 

The robustness of SAM~\cite{kirillov2023segment} and SAM 2~\cite{ravi2024sam} can vary depending on the type of corruption, but generally, the performance tends to decline noticeably. Particularly, for SAM~\cite{kirillov2023segment}, \textit{JPEG Compression} and \textit{Gaussian Noise} exert the most pronounced effects on segmentation performance, whereas \textit{Brightness} has minimal impact. When subjected to identical levels and types of corruption, SAM 2 demonstrates less performance degradation compared to SAM, indicating that SAM 2 exhibits greater robustness than SAM. However, SAM 2 and SAM maintained consistency in the most and least affected corruption types. Specifically, \textit{Gaussian Noise}, \textit{JPEG Compression} and \textit{Zoom} significantly affect the segmentation performance of SAM 2, whereas \textit{Brightness} has a minor influence. SAM 2-Video Segmentation has the worst robustness. This is reflected in the fact that its performance degrades more than SAM-Image Segmentation and SAM 2-Image Segmentation when faced with corrupted images. Especially when faced with corrupted images with a severity level of 5, its performance drops sharply. This may be because point prompts, although more convenient, make the model unable to track tools well and cope with complex variations. Figure~\ref{fig:corrupt} displays an initial frame alongside several altered versions under severity level 5. It is evident from the images that SAM~\cite{kirillov2023segment} and SAM 2~\cite{ravi2024sam} experience significant performance degradation across most scenarios.

\section{Conclusion}\label{sec:conclusion}

In this empirical study, we build upon prior work~\cite{wang2023sam} by further investigating the zero-shot capabilities and data corruption robustness of SAM~2~\cite{ravi2024sam} in semantic segmentation for robot-assisted surgery. Our analysis is primarily based on two types of prompts: single point and bounding box. Under the bounding box prompt, SAM~2~\cite{ravi2024sam} maintains the exceptional performance observed in SAM~\cite{kirillov2023segment}, achieving SOTA results with slight improvements over SAM~\cite{kirillov2023segment}. In contrast, the 1-point prompt results from SAM~\cite{kirillov2023segment} exhibited subpar performance, making precise segmentation of surgical instruments challenging. However, SAM~2~\cite{ravi2024sam} demonstrates high performance with the 1-point prompt, producing satisfactory inference results compared to previous work – even with this simple prompting approach, which significantly advances downstream applications. 
Furthermore, in video segmentation, although we only employ a single point from the initial frame as the prompt, SAM~2~\cite{ravi2024sam} exhibits better results than SAM~\cite{kirillov2023segment}'s image segmentation, which utilizes the 1-point prompt for each frame. When encountering various types of image quality corruption, SAM~2~\cite{ravi2024sam} also demonstrates less performance degradation compared to SAM~\cite{kirillov2023segment}, highlighting its exceptional ability to handle complex scenarios including occlusions, noise, blur, and other challenges in downstream tasks for robotic-assisted surgery.

Nevertheless, SAM~2~\cite{ravi2024sam} still faces certain limitations, such as suboptimal segmentation performance in some edge cases or regions. Future work may focus on the development of automated, prompt-free methods for automated segmentation in surgical settings. Exploring ways for the SAM model to better comprehend the textual representations corresponding to images would also be an interesting avenue of research.

\subsubsection{Acknowledgements.}

This work was supported by Hong Kong Research Grants Council (RGC) Collaborative Research Fund (CRF C4026-21GF), General Research Fund (GRF 14203323, GRF 14216022, and GRF 14211420),  NSFC/RGC Joint Research Scheme N\_CUHK420/22; Shenzhen-Hong Kong-Macau Technology Research Programme (Type C) STIC Grant 202108233000303.

\bibliographystyle{splncs04}
\bibliography{references}

\end{document}